\pdfoutput=1

\documentclass[11pt]{article}

\usepackage[]{acl}

\usepackage{multicol} 
\usepackage{amssymb}
\usepackage{fontawesome}
\usepackage{booktabs}

\usepackage{times}
\usepackage{latexsym}
\usepackage{amsmath}
\usepackage[T1]{fontenc}
\usepackage[utf8]{inputenc}
\usepackage{algorithm}
\usepackage{algpseudocode}
\usepackage{microtype}
\usepackage{inconsolata}
\usepackage{graphicx}
\usepackage{pifont}
\usepackage{mathrsfs}
\usepackage{subcaption}
\usepackage{pgfplots}
\pgfplotsset{compat=newest} 
\usepackage{tabularx}  
\usepackage{multirow}   
\usepackage[most]{tcolorbox}
\usepackage{tikz}
\usepackage{lipsum} 

\usepackage{listings}
\usepackage{siunitx} 
\usepackage{authblk}

\tcbuselibrary{breakable}
\usetikzlibrary{patterns}

\lstdefinestyle{jsonstyle}{
    language=Python,
    basicstyle=\small\ttfamily,
    keywordstyle=\color{blue},
    stringstyle=\color{blue}, 
    commentstyle=\color{green},
    morestring=[b]",
    tabsize=2,
    showstringspaces=false
}

\title{From Model-centered to Human-Centered: Revision Distance as a Metric for Text Evaluation in LLMs-based Applications}

\author{

    Yongqiang Ma$^{1,2,}$\thanks{This work was done when Yongqiang Ma interned at Alibaba.}, Lizhi Qing$^2$, Jiawei Liu$^1$, Yangyang Kang$^2$, \\Yue Zhang$^2$, Wei Lu$^1$, Xiaozhong Liu$^3$, Qikai Cheng$^1$\hspace{0.3cm}\\
    $^1$School of Information Management, Wuhan University, China \hspace{0.5cm} \\$^2$Institute for Intelligent Computing, Alibaba Group, China \hspace{0.5cm}\\
    $^3$Worcester Polytechnic Institute, USA\\
\texttt{\{mayongqiang,laujames2017,weilu\}@whu.edu.cn}\\
\texttt{\{yekai.qlz,shiyu.zy,yangyang.kangyy\}@alibaba-inc.com}\\
\texttt{xliu14@wpi.edu, chengqikai0806@163.com }
}

\begin{document}
\maketitle
\begin{abstract}

Evaluating large language models (LLMs) is fundamental, particularly in the context of practical applications.  
Conventional evaluation methods, typically designed primarily for LLM development, yield numerical scores that ignore the user experience. Therefore, our study shifts the focus from model-centered to human-centered evaluation in the context of AI-powered writing assistance applications. 
Our proposed metric, termed ``Revision Distance,'' utilizes LLMs to suggest revision edits that mimic the human writing process. It is determined by counting the revision edits generated by LLMs. Benefiting from the generated revision edit details, our metric can provide a self-explained text evaluation result in a human-understandable manner beyond the context-independent score. 
Our results show that for the easy-writing task, ``Revision Distance'' is consistent with established metrics (ROUGE, Bert-score, and GPT-score), but offers more insightful, detailed feedback and better distinguishes between texts. 
Moreover, in the context of challenging academic writing tasks, our metric still delivers reliable evaluations where other metrics tend to struggle. Furthermore, our metric also holds significant potential for scenarios lacking reference texts.

\end{abstract}

\section{Introduction}

\begin{quote}
    You can’t manage what you can’t measure well.—\citealt{CRUZCAZARES20131239} 
\end{quote}


With the continuous development of large language models (LLMs) such as ChatGPT\footnote{\url{https://openai.com/blog/chatgpt}}, GPT-4\citep{openaiGPT4TechnicalReport}, and Llama\citep{touvron2023llama}, a plethora of application research and development work based on LLMs has emerged.

During the model training phase, the main focus is optimizing the model's loss in an isolated environment. However, LLM-based applications should be human-centered, prioritizing user experience and utility. This raises a key question: How do we evaluate LLM-based applications from a human-centric perspective?

\begin{figure}[t]
    \centering
    \includegraphics[scale=0.65]{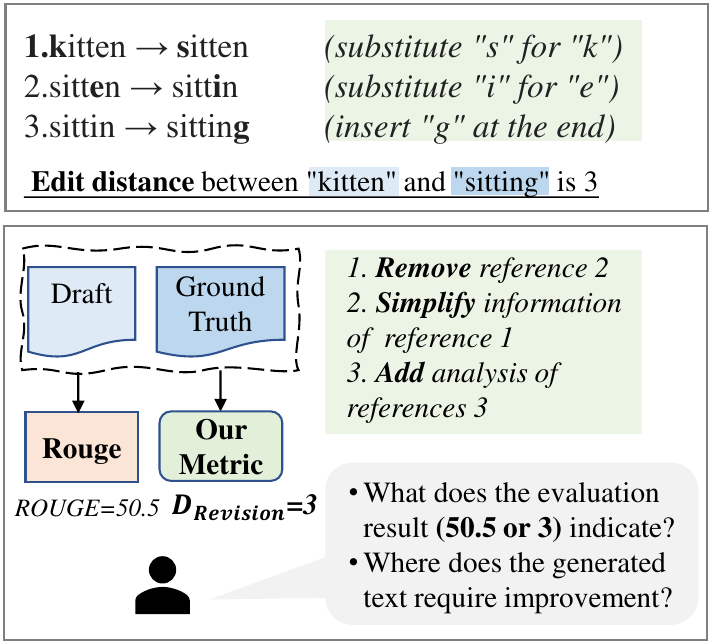}
    \caption{Inspired by the classical edit distance metric, our ``Revision Distance'' $\mathbf{D}_{Revision}$ can offer a more human-centered and nuanced metric for text evaluation. As illustrated, the $\mathbf{D}_{Revision}(Draft, GroudTruth)$ can provide a more transparent evaluation result, benefiting from the generated revision edit details.}
    \label{fig:enter-label}
\end{figure}

Imagining the scenario where developers employ automatic evaluation metrics \citep{lin-2004-rouge,papineni-etal-2002-bleu,bert-score,zhao-etal-2019-moverscore} like ROUGE\citep{lin-2004-rouge} to evaluate LLM-generated text for writing assistance debugging. Whereas ROUGE only provides a high-level evaluation score to measure textual surface similarity. 
Since disregarding end-users, the evaluation result is inadequate and misaligns with user needs and preferences.
To address this gap, we explore alternative human-centered evaluation metrics, putting the user at the forefront of our evaluation. 

This paper focuses on the prevalent application scenario for LLMs, specifically, the LLM-powered writing assistant in easy-writing scenarios and challenge-writing scenarios from email, and letter writing to academic writing \footnote{We use the ``Related Work'' section Generation (RWG) \citep{liu-etal-2023-causal,chen-etal-2021-capturing} as the testbed for academic writing, which requires heavy knowledge reasoning work and complex concept understanding ability.}.
During the AI-human collaborative writing process, AI-generated text often requires extended revisions. 
Additionally, recent studies suggest that LLMs can produce human-like behavior, such as providing human preferences feedback \citep{bai2022constitutional,lee2023rlaif}, conducting text quality evaluation \citep{chiang-lee-2023-large,fu2023gptscore}. Therefore, we assume that the LLM can be a proxy user for generating revision edits, aligning with actual human editing behaviors. 
Drawing from these insights, our proposed metric, $Revision Distance$, incorporates the iterative process of user-driven text revision. It quantifies the number of edits a user must take to an LLM-generated text to achieve a predefined quality threshold. 

In the reference-based evaluation setting, we compared our metric with ROUGE, BERT-Score, and GPT-Score across two writing tasks: the easy-writing task and the challenge-writing task. For each task, we sample texts from two models to form a comparison group. Then we apply text evaluation metrics to assess the text quality. 
(1) For the easy-writing task, we find that our metric consistently aligns with baseline metrics, supporting the intuition that a stronger model should produce texts with superior evaluation scores. 
(2) For more challenging tasks, our metrics can still provide stable and reliable evaluation results even if most of the baseline indicators encounter different issues.

In reference-free scenarios, the ``Revision Distance'' metric aligns closely with human judgment in approximately 76\% of cases in the dataset from ultrafeedback dataset \citep{notus2023}. Furthermore, by categorizing the types of edits made, our metric provides a more fine-grained analysis than those metrics that only yield scores.

The contributions are listed as follows: 1) We highlight the significance of the end-user's perspective in the text evaluation in the context of LLM-power writing assistant. 2) Aligning with real-world human editing behaviors, we propose a human-centered text evaluation metric, which provides a self-explain and fine-grained insight for developers and end-users. 3) Based on broad and various test tasks, we conduct an experiment to demonstrate the utility of our proposed human-centered metrics.

\section{Related Work}

The text evaluation methods can be categorized into human evaluation and automated approaches. Human evaluation is widely recognized as the most natural way to evaluate the quality of a given text, which often involves human annotators and qualitative analyses \citep{clark-etal-2021-thats,belz-etal-2023-non}. This method is often expensive and time-consuming work and requires extensive domain expertise for domain-specific scenarios. 
On the other hand, current automated evaluation methods tend to generate a comprehensive score that is facilitated in comparing new models with established state-of-the-art approaches. These include metrics such as ROUGE \citep{lin-2004-rouge}, BLEU \citep{papineni-etal-2002-bleu}, BERTScore \citep{bert-score}, MoverScore \citep{zhao-etal-2019-moverscore}, BARTScore \citep{BARTScore}, and DiscoScore \citep{zhao-etal-2023-discoscore}  typically compute a similarity (or dissimilarity) score between a model-generated text and a reference text. 

Large language models have been adeptly utilized for roles such as aiding in data annotation \citep{li-etal-2023-coannotating} and delivering feedback that mirrors human preferences \citep{bai2022constitutional,lee2023rlaif,pang2023language}. 
For the evaluation stage, \citet{chiang-lee-2023-large} found that the LLM evaluation is consistent with the human evaluation results. The GPTScore \citep{fu2023gptscore} has been proposed to score the model-generated text. Similarly, \citep{jain-etal-2023-multi} also studied the efficacy of LLMs as multi-dimensional evaluators. 

In conclusion, current metrics tend to yield a comprehensive score that detaches the task context for model development and optimization. However, the ultimate application of LLMs is human-centered, prioritizing the user experience and utility. 
Consequently, a context-independent numerical score is insufficient in LLM application scenarios. 
Our proposed metric shifts the text evaluation to a human-centered perspective, which incorporates the iterative process of user-driven text revision.

\section{Revision Distance}

\begin{figure*}[htbp]
\centering
\includegraphics[scale=0.63]{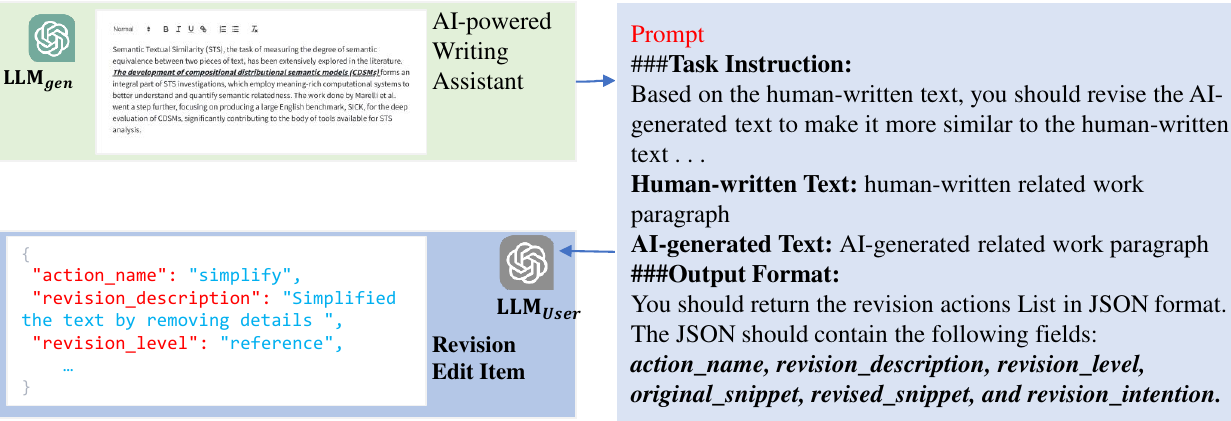}

\caption{The evaluation flow of ``Revision Distance''. We require the $LLM_{User}$ to produce results in JSON format with detailed information, In this work, we primarily use the \textbf{action\_name} to analyze the revisions.}
\label{fig:boxrevise}
\end{figure*}

As depicted in Figure \ref{fig:boxrevise}, we frame the context of AI-powered writing assistance, with LLMs serving dual functions: as the proxy of the user ($LLM_{User}$) and as the generator ($LLM_{gen}$). The $LLM_{gen}$ acts as the pivotal component of the writing assistant application. 
Given the user input content, the $LLM_{gen}$ generates a draft $Y_{draft}$, such as emails, letters, articles, and ``Related Work'' sections. The $\mathbf{D}_{Revision}$ quantifies the number of edits from $LLM_{User}(Y_{draft},Y)$.

For the reference-based evaluation setting, we utilize the human-written text or ChatGPT output as the ground truth. The $LLM_{User}$ is designed to produce structured revision edits, improving the consistency of the $Y_{Draft}$ to the ground-truth text $Y$. In scenarios where no ground truth text is available, we require the $LLM_{User}$  to refine the given text towards an ideal form, as envisioned by the $LLM_{User}$ itself\footnote{This ideal version is not explicitly generated but rather serves as an implicit standard within the revision edits generation prompt.}.
These revision edits are produced to improve $Y_{draft}$ to closer align with the ideal version, which mimics the revision process of human writers. 

\section{Results and Discussion}

\subsection{Evaluation for Reference-based Setting}

\subsubsection{Task and Dataset}

To validate the utility of our proposed metric, we have constructed two distinct datasets to address both the easy-writing task and the challenge-writing task. The challenge-writing task refers to the scenario that requires heavy knowledge reasoning and complex concept understanding. For the easy-writing task, we use the task of emails, letters, and articles generation as a testbed.  For the challenge-writing task, we employ academic writing as the testbed. The test dataset details in this evaluation setting are described in Appendix \ref{dataset}.

\subsubsection{Text Generation Models}

To assess the discriminative capacity of our revision distance metric, we designed strong and weak writing applications. The terms ``strong'' and ``weak''  refer to the generation ability of utilized LLM, as detailed in Table \ref{tab:models}. (1) For the easy-writing task, we employ two Mistral-series models \citep{jiang2023mistral}; (2) For the challenge-writing task, we employ GPT-4 and its variant \footnote{The models employed in both tasks are detailed in Appendix \ref{mistral} and Appendix \ref{sec:cot}, respectively.}.

\begin{table}[ht]
    \centering
    \begin{tabular}{ccc}
    \toprule
        Task Level & Weak \includegraphics[scale=0.022]{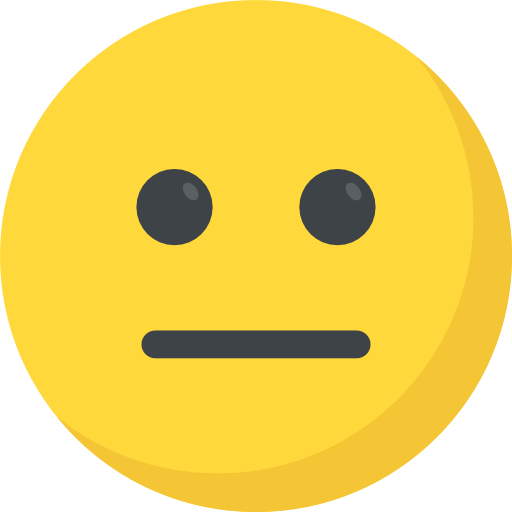}  &  Strong \includegraphics[scale=0.022]{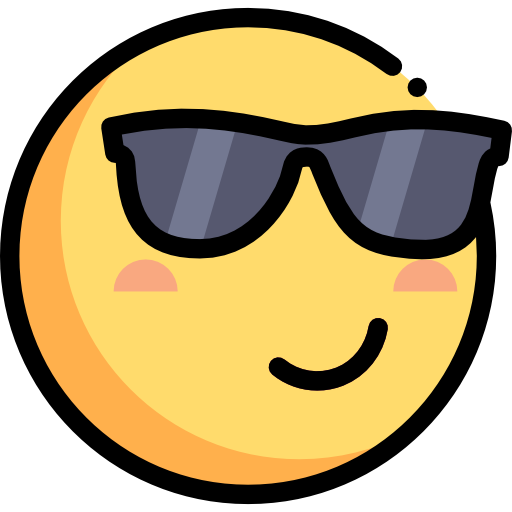} \\
        \midrule
        Easy & Mistral-7B & Mixtral-8x7B\\
        Challenge  & vanilla GPT-4 & CoT-based GPT-4  \\
        \bottomrule
    \end{tabular}
    \caption{The employed models for both writing tasks.}
    \label{tab:models}
\end{table}

\subsubsection{Result Analysis}

\begin{table*}[htbp]
  \centering
  \begin{tabular}{
    l
    S[table-format=2.2, table-column-width=15mm, table-align-text-post=true]
    @{\,}l
    @{\,}r
    S[table-format=2.2, table-column-width=15mm, table-align-text-post=false]
    @{\,}l
    @{\,}r
    S[table-format=2.2, table-column-width=15mm, table-align-text-post=true]
    l
    r
    S[table-format=2.2, table-column-width=15mm, table-align-text-post=false]
    @{\,}l
    @{\,}r
  }
    \toprule
    & \multicolumn{6}{c}{Easy-writing Task} & \multicolumn{6}{c}{Challenge-writing Task} \\
    \cmidrule(lr){2-7} \cmidrule(lr){8-13}
    
    Metrics & \multicolumn{3}{c}{\includegraphics[scale=0.022]{confused.png} Mistral-7b} & \multicolumn{3}{c}{\includegraphics[scale=0.022]{cool.png}Mixtral-8x7b} & \multicolumn{3}{c}{\includegraphics[scale=0.022]{confused.png} Vanilla GPT-4} & \multicolumn{3}{c}{\includegraphics[scale=0.022]{cool.png}CoT GPT-4} \\
    \cmidrule(lr){1-1} \cmidrule(lr){2-7} \cmidrule(lr){8-13}
    
    Rouge 1 $\uparrow$ & 50.53 & & & 51.65 & \textcolor{green}{\faCaretUp} & 2.2\% & 31.62 & & & 29.78 & \textcolor{red}{\faCaretDown} & -6.2\% \\
    Rouge 2 $\uparrow$ & 19.52 & & & 22.06 & \textcolor{green}{\faCaretUp} & 11.5\% & 7.64  & & & 6.86  & \textcolor{red}{\faCaretDown} & -11.4\% \\
    Rouge L $\uparrow$ & 26.74 & & & 29.21 & \textcolor{green}{\faCaretUp} & 8.5\% & 15.09 & & & 15.59 & \textcolor{green}{\faCaretUp} & 3.2\% \\
    Bert-Score $\uparrow$ & {--} & & & {--} & & {--} & 57.36 & & & 56.36 & \textcolor{red}{\faCaretDown} & -1.8\% \\
    GPT-Score $\uparrow$ & 90.56 & & & 88.63 & \textcolor{red}{\faCaretDown} & -2.2\% & 87.67 & & & 87.47 & \textcolor{red}{\faCaretDown} & -0.2\% \\
    \textbf{$\mathbf{D}_{Revision}$} $\downarrow$& 3.20 & & & 2.79  & \textcolor{green}{\faCaretUp} & \textbf{14.7\%} & 3.94  & & & 3.73  & \textcolor{green}{\faCaretUp} & \textbf{5.3\%} \\
    \bottomrule
  \end{tabular}
  \caption{The symbols \textcolor{green}{\faCaretUp} and \textcolor{red}{\faCaretDown} indicate directional changes in performance as delivered by evaluation metrics. Specifically, \textcolor{green}{\faCaretUp} and \textcolor{red}{\faCaretDown} denote performance improvement and decline, respectively, from weaker to stronger models. For the easy-writing task, our $\mathbf{D}_{Revision}$ aligns well with other evaluation measures, which shows the utility of our metric. For the more challenging writing task, it offers stable evaluations and better distinguishes model quality, whereas other metrics struggle. The limited input length of Bert-Score, capped at 512 tokens, precluded its use in the easy-writing task where many texts exceeded this limit.
  }

  \label{tab:eval_result}

\end{table*}

As shown in Table \ref{tab:eval_result}, our metric shows utility for easy and challenging writing tasks. Different from other metrics, smaller $\mathbf{D}_{Revision}$ indicate better text quality. 
To assess the metric's ability to differentiate between models, we calculate the relative change rate from the ``Weak'' model to the ``Strong'' model. 
Notably, existing metrics have reached saturation for the easy-writing tasks, exhibiting a limited relative change rate regarding the performance of distinct models. Conversely, our metric demonstrates better efficacy in discerning the nuanced capabilities of diverse models. 
It's observed that $\mathbf{D}_{Revision}$ yields a larger change rate, highlighting the enhanced discriminative capacity of our metric.

Additionally, for the complex academic writing task, we conducted a human evaluation\footnote{We selected 20 paragraphs from both methods for expert analysis. Five AI field specialists assessed the LLM-generated content, focusing on content quality, structural coherence, and argumentative strength.}. Based on the evaluation results, we categorized texts as ``Chosen'' or ``Rejected.'' Our $\mathbf{D}_{Revision}$ metric aligns with human preferences, indicating superior text quality with fewer revisions for ``Chosen Texts.'' In contrast, the ROUGE metric often misaligns with human judgments, erroneously assigning higher scores to ``Rejected Texts.''

\subsection{Evaluation for Reference-free Setting}

To demonstrate the performance of our evaluation method in scenarios, where ground truth is unavailable, we extracted 41 cases related to writing tasks from the UltraFeedback dataset\citep{notus2023}. Each case contains a chosen response and a rejected response. 

When applied to the selected cases, our ``Revision Distance'' metric aligns with human judgments in 76\% of instances, indicating that chosen responses typically necessitated fewer revisions.
\subsection{Qualitative Analysis}

Based on the analysis of revision edit details, we classify the revision actions into three categories: (1) Reference Order Revision, (2) Reference Comparison Revision, and (3) Reference Description Revision. The description of three categories is shown in Appendix \ref{cat_rev}. 

For complex writing tasks, the challenge often lies in knowledge reasoning of concepts. CoT prompting can dramatically improve the multi-step reasoning abilities of LLMs \citep{wang-etal-2023-towards}. As shown in Table \ref{tab:revise_cnt}, CoT-based GPT-4 can provide text with fewer revisions related to Order and Comparison issues in ``Related work'' writing tasks. 
The improvements can be attributed to the enhanced knowledge reasoning capabilities of the CoT-based method. There also exists a slight decline in the reference description dimension. 
In conclusion, the fine-grained analysis revision edits can provide insightful feedback for future model improvement.

\begin{table}
    \centering
    \begin{tabular}{ccccc}
    \toprule
        Model Type & Order & Comp  & Desc & Total \\
    \midrule
    Vanilla GPT-4   & 0.80  & 0.84 & 2.29 &  3.93\\
    CoT GPT-4     & 0.67 & 0.71 & 2.36 & 3.73\\
    \bottomrule
    \end{tabular}
    \caption{The result of revision edits analysis. }
    \label{tab:revise_cnt}
\end{table}

\section{Conclusion}

With the rapid advancement of LLM-based applications, the pivotal question arises:
``how can we evaluate LLM-based applications from a human-centered perspective?'' 
Existing evaluation metrics, typically used for model development, merely yield a context-independent numerical score, lacking user relevance. 
Our research shifts text evaluation from a predominantly model-centered perspective to a human-centered one. 

Using the LLM-powered writing assistant as a test scenario, we take a comprehensive experiment on diverse writing tasks to validate the effectiveness and reliability of our ``Revision Distance'' metric. 
This metric converts text evaluation into contextualized text revisions, clearly highlighting textual discrepancies and offering users a detailed, transparent rationale for the scores. 
Our findings demonstrate the metric's applicability and dependability in both reference-based and reference-free contexts.

\section*{Limitations}

This paper introduces a metric that leverages GPT-4, specifically applied to evaluating  LLM-powered writing assistants. However, the computational and financial costs of using GPT-4 are considerable. Exploring the use of a smaller, specialized model to generate initial edits could reduce costs and improve efficiency. 

LLMs have a wide array of applications, and for this study, we have chosen the ``Related Work'' section generation task as a testbed for challenging writing scenarios. As a knowledge-intensive cognitive task, writing the ``Related Work'' section requires writers to integrate multi-source knowledge into the manuscript. Therefore, writing a comprehensive ``Related Work'' section is a labor-intensive and time-consuming endeavor. Future studies could explore the application of our metric in longer text generation tasks, such as code generation and scientific reports, to validate its effectiveness and applicability across different domains.

In this study,  each generated revision item is assigned equal weight. Future research should focus on developing a dynamic revision edit weighting method to evaluate textual differences more finely. 

\section*{Ethics Statement}
In this paper, we propose a new automatic evaluation metric $Revision Distance$ to evaluate the LLM-generated text in an AI-power writing assistant setting. The positive impact of Revision Distance is that it can provide a more nuanced and self-explain representation of the quality of LLM-generated text. Notably, our metric is human-centered and transparent, which can help demystify the evaluation process for LLM-generated text, making it more accessible to a wider user, including those who are not experts in AI. 
The negative impact is that over-reliance on $Revision Distance$ might lead to the overlooking of qualitative aspects of text generation that are harder to quantify, such as creativity. Additionally, if the reference texts within $Revision Distance$ are biased or of low quality, this could amplify the biases in the LLMs-generated text.

\section*{Acknowledgements}
\bibliography{custom}

\begin{thebibliography}{26}
\expandafter\ifx\csname natexlab\endcsname\relax\def\natexlab#1{#1}\fi

\bibitem[{Bai et~al.(2022)Bai, Kadavath, Kundu, Askell, Kernion, Jones, Chen, Goldie, Mirhoseini, McKinnon, Chen, Olsson, Olah, Hernandez, Drain, Ganguli, Li, Tran-Johnson, Perez, Kerr, Mueller, Ladish, Landau, Ndousse, Lukosuite, Lovitt, Sellitto, Elhage, Schiefer, Mercado, DasSarma, Lasenby, Larson, Ringer, Johnston, Kravec, Showk, Fort, Lanham, Telleen-Lawton, Conerly, Henighan, Hume, Bowman, Hatfield-Dodds, Mann, Amodei, Joseph, McCandlish, Brown, and Kaplan}]{bai2022constitutional}
Yuntao Bai, Saurav Kadavath, Sandipan Kundu, Amanda Askell, Jackson Kernion, Andy Jones, Anna Chen, Anna Goldie, Azalia Mirhoseini, Cameron McKinnon, Carol Chen, Catherine Olsson, Christopher Olah, Danny Hernandez, Dawn Drain, Deep Ganguli, Dustin Li, Eli Tran-Johnson, Ethan Perez, Jamie Kerr, Jared Mueller, Jeffrey Ladish, Joshua Landau, Kamal Ndousse, Kamile Lukosuite, Liane Lovitt, Michael Sellitto, Nelson Elhage, Nicholas Schiefer, Noemi Mercado, Nova DasSarma, Robert Lasenby, Robin Larson, Sam Ringer, Scott Johnston, Shauna Kravec, Sheer~El Showk, Stanislav Fort, Tamera Lanham, Timothy Telleen-Lawton, Tom Conerly, Tom Henighan, Tristan Hume, Samuel~R. Bowman, Zac Hatfield-Dodds, Ben Mann, Dario Amodei, Nicholas Joseph, Sam McCandlish, Tom Brown, and Jared Kaplan. 2022.
\newblock Constitutional ai: Harmlessness from ai feedback.
\newblock \emph{arXiv preprint arXiv:2212.08073}.

\bibitem[{Bartolome et~al.(2023)Bartolome, Martin, and Vila}]{notus2023}
Alvaro Bartolome, Gabriel Martin, and Daniel Vila. 2023.
\newblock Notus.
\newblock \url{https://github.com/argilla-io/notus}.

\bibitem[{Belz et~al.(2023)Belz, Thomson, Reiter, and Mille}]{belz-etal-2023-non}
Anya Belz, Craig Thomson, Ehud Reiter, and Simon Mille. 2023.
\newblock \href {https://doi.org/10.18653/v1/2023.findings-acl.226} {Non-repeatable experiments and non-reproducible results: The reproducibility crisis in human evaluation in {NLP}}.
\newblock In \emph{Findings of the Association for Computational Linguistics: ACL 2023}, pages 3676--3687, Toronto, Canada. Association for Computational Linguistics.

\bibitem[{Chen et~al.(2021)Chen, Alamro, Li, Gao, Zhang, Zhao, and Yan}]{chen-etal-2021-capturing}
Xiuying Chen, Hind Alamro, Mingzhe Li, Shen Gao, Xiangliang Zhang, Dongyan Zhao, and Rui Yan. 2021.
\newblock \href {https://doi.org/10.18653/v1/2021.acl-long.473} {Capturing relations between scientific papers: An abstractive model for related work section generation}.
\newblock In \emph{Proceedings of the 59th Annual Meeting of the Association for Computational Linguistics and the 11th International Joint Conference on Natural Language Processing (Volume 1: Long Papers)}, pages 6068--6077, Online. Association for Computational Linguistics.

\bibitem[{Chiang and Lee(2023)}]{chiang-lee-2023-large}
Cheng-Han Chiang and Hung-yi Lee. 2023.
\newblock \href {https://doi.org/10.18653/v1/2023.acl-long.870} {Can large language models be an alternative to human evaluations?}
\newblock In \emph{Proceedings of the 61st Annual Meeting of the Association for Computational Linguistics (Volume 1: Long Papers)}, pages 15607--15631, Toronto, Canada. Association for Computational Linguistics.

\bibitem[{Clark et~al.(2021)Clark, August, Serrano, Haduong, Gururangan, and Smith}]{clark-etal-2021-thats}
Elizabeth Clark, Tal August, Sofia Serrano, Nikita Haduong, Suchin Gururangan, and Noah~A. Smith. 2021.
\newblock \href {https://doi.org/10.18653/v1/2021.acl-long.565} {All that{'}s {`}human{'} is not gold: Evaluating human evaluation of generated text}.
\newblock In \emph{Proceedings of the 59th Annual Meeting of the Association for Computational Linguistics and the 11th International Joint Conference on Natural Language Processing (Volume 1: Long Papers)}, pages 7282--7296, Online. Association for Computational Linguistics.

\bibitem[{Cruz-Cázares et~al.(2013)Cruz-Cázares, Bayona-Sáez, and García-Marco}]{CRUZCAZARES20131239}
Claudio Cruz-Cázares, Cristina Bayona-Sáez, and Teresa García-Marco. 2013.
\newblock \href {https://doi.org/https://doi.org/10.1016/j.respol.2013.03.012} {You can’t manage right what you can’t measure well: Technological innovation efficiency}.
\newblock \emph{Research Policy}, 42(6):1239--1250.

\bibitem[{Fu et~al.(2023)Fu, Ng, Jiang, and Liu}]{fu2023gptscore}
Jinlan Fu, See-Kiong Ng, Zhengbao Jiang, and Pengfei Liu. 2023.
\newblock Gptscore: Evaluate as you desire.
\newblock \emph{arXiv preprint arXiv:2302.04166}.

\bibitem[{Jain et~al.(2023)Jain, Keshava, Mysore~Sathyendra, Fernandes, Liu, Neubig, and Zhou}]{jain-etal-2023-multi}
Sameer Jain, Vaishakh Keshava, Swarnashree Mysore~Sathyendra, Patrick Fernandes, Pengfei Liu, Graham Neubig, and Chunting Zhou. 2023.
\newblock \href {https://doi.org/10.18653/v1/2023.findings-acl.537} {Multi-dimensional evaluation of text summarization with in-context learning}.
\newblock In \emph{Findings of the Association for Computational Linguistics: ACL 2023}, pages 8487--8495, Toronto, Canada. Association for Computational Linguistics.

\bibitem[{Jiang et~al.(2023)Jiang, Sablayrolles, Mensch, Bamford, Chaplot, de~las Casas, Bressand, Lengyel, Lample, Saulnier, Lavaud, Lachaux, Stock, Scao, Lavril, Wang, Lacroix, and Sayed}]{jiang2023mistral}
Albert~Q. Jiang, Alexandre Sablayrolles, Arthur Mensch, Chris Bamford, Devendra~Singh Chaplot, Diego de~las Casas, Florian Bressand, Gianna Lengyel, Guillaume Lample, Lucile Saulnier, Lélio~Renard Lavaud, Marie-Anne Lachaux, Pierre Stock, Teven~Le Scao, Thibaut Lavril, Thomas Wang, Timothée Lacroix, and William~El Sayed. 2023.
\newblock Mistral 7b.
\newblock \emph{arXiv preprint arXiv:2310.06825}.

\bibitem[{Lee et~al.(2023)Lee, Phatale, Mansoor, Mesnard, Ferret, Lu, Bishop, Hall, Carbune, Rastogi, and Prakash}]{lee2023rlaif}
Harrison Lee, Samrat Phatale, Hassan Mansoor, Thomas Mesnard, Johan Ferret, Kellie Lu, Colton Bishop, Ethan Hall, Victor Carbune, Abhinav Rastogi, and Sushant Prakash. 2023.
\newblock Rlaif: Scaling reinforcement learning from human feedback with ai feedback.
\newblock \emph{arXiv preprint arXiv:2309.00267}.

\bibitem[{Li et~al.(2023)Li, Shi, Ziems, Kan, Chen, Liu, and Yang}]{li-etal-2023-coannotating}
Minzhi Li, Taiwei Shi, Caleb Ziems, Min-Yen Kan, Nancy Chen, Zhengyuan Liu, and Diyi Yang. 2023.
\newblock \href {https://aclanthology.org/2023.emnlp-main.92} {{C}o{A}nnotating: Uncertainty-guided work allocation between human and large language models for data annotation}.
\newblock In \emph{Proceedings of the 2023 Conference on Empirical Methods in Natural Language Processing}, pages 1487--1505, Singapore. Association for Computational Linguistics.

\bibitem[{Lin(2004)}]{lin-2004-rouge}
Chin-Yew Lin. 2004.
\newblock \href {https://aclanthology.org/W04-1013} {{ROUGE}: A package for automatic evaluation of summaries}.
\newblock In \emph{Text Summarization Branches Out}, pages 74--81, Barcelona, Spain. Association for Computational Linguistics.

\bibitem[{Liu et~al.(2023)Liu, Zhang, Shi, Naseem, Wang, Hu, and Tsang}]{liu-etal-2023-causal}
Jiachang Liu, Qi~Zhang, Chongyang Shi, Usman Naseem, Shoujin Wang, Liang Hu, and Ivor Tsang. 2023.
\newblock \href {https://doi.org/10.18653/v1/2023.findings-emnlp.141} {Causal intervention for abstractive related work generation}.
\newblock In \emph{Findings of the Association for Computational Linguistics: EMNLP 2023}, pages 2148--2159, Singapore. Association for Computational Linguistics.

\bibitem[{OpenAI()}]{openaiGPT4TechnicalReport}
OpenAI.
\newblock {GPT}-4 technical report.
\newblock Technical report, OpenAI.

\bibitem[{Pang et~al.(2023)Pang, Wang, Li, Chen, Xu, Zhang, and Yu}]{pang2023language}
Jing-Cheng Pang, Pengyuan Wang, Kaiyuan Li, Xiong-Hui Chen, Jiacheng Xu, Zongzhang Zhang, and Yang Yu. 2023.
\newblock Language model self-improvement by reinforcement learning contemplation.
\newblock \emph{arXiv preprint arXiv:2305.14483}.

\bibitem[{Papineni et~al.(2002)Papineni, Roukos, Ward, and Zhu}]{papineni-etal-2002-bleu}
Kishore Papineni, Salim Roukos, Todd Ward, and Wei-Jing Zhu. 2002.
\newblock \href {https://doi.org/10.3115/1073083.1073135} {{B}leu: a method for automatic evaluation of machine translation}.
\newblock In \emph{Proceedings of the 40th Annual Meeting of the Association for Computational Linguistics}, pages 311--318, Philadelphia, Pennsylvania, USA. Association for Computational Linguistics.

\bibitem[{Rohatgi et~al.(2023)Rohatgi, Qin, Aw, Unnithan, and Kan}]{rohatgi2023acl}
Shaurya Rohatgi, Yanxia Qin, Benjamin Aw, Niranjana Unnithan, and Min-Yen Kan. 2023.
\newblock The acl ocl corpus: advancing open science in computational linguistics.
\newblock \emph{arXiv preprint arXiv:2305.14996}.

\bibitem[{Sun et~al.(2023)Sun, Wang, Tay, Yang, and Zhou}]{sun2023recitationaugmented}
Zhiqing Sun, Xuezhi Wang, Yi~Tay, Yiming Yang, and Denny Zhou. 2023.
\newblock \href {https://openreview.net/forum?id=-cqvvvb-NkI} {Recitation-augmented language models}.
\newblock In \emph{International Conference on Learning Representations}.

\bibitem[{Touvron et~al.(2023)Touvron, Martin, Stone, Albert, Almahairi, Babaei, Bashlykov, Batra, Bhargava, Bhosale et~al.}]{touvron2023llama}
Hugo Touvron, Louis Martin, Kevin Stone, Peter Albert, Amjad Almahairi, Yasmine Babaei, Nikolay Bashlykov, Soumya Batra, Prajjwal Bhargava, Shruti Bhosale, et~al. 2023.
\newblock Llama 2: {{Open}} foundation and fine-tuned chat models.
\newblock \emph{arXiv preprint arXiv:2307.09288}.

\bibitem[{Wang et~al.(2023)Wang, Min, Deng, Shen, Wu, Zettlemoyer, and Sun}]{wang-etal-2023-towards}
Boshi Wang, Sewon Min, Xiang Deng, Jiaming Shen, You Wu, Luke Zettlemoyer, and Huan Sun. 2023.
\newblock \href {https://doi.org/10.18653/v1/2023.acl-long.153} {Towards understanding chain-of-thought prompting: An empirical study of what matters}.
\newblock In \emph{Proceedings of the 61st Annual Meeting of the Association for Computational Linguistics (Volume 1: Long Papers)}, pages 2717--2739, Toronto, Canada. Association for Computational Linguistics.

\bibitem[{Yuan et~al.(2021)Yuan, Neubig, and Liu}]{BARTScore}
Weizhe Yuan, Graham Neubig, and Pengfei Liu. 2021.
\newblock \href {https://proceedings.neurips.cc/paper_files/paper/2021/file/e4d2b6e6fdeca3e60e0f1a62fee3d9dd-Paper.pdf} {Bartscore: Evaluating generated text as text generation}.
\newblock In \emph{Advances in Neural Information Processing Systems}, volume~34, pages 27263--27277. Curran Associates, Inc.

\bibitem[{Zhang et~al.(2020)Zhang, Kishore, Wu, Weinberger, and Artzi}]{bert-score}
Tianyi Zhang, Varsha Kishore, Felix Wu, Kilian~Q. Weinberger, and Yoav Artzi. 2020.
\newblock Bertscore: Evaluating text generation with bert.
\newblock In \emph{International Conference on Learning Representations}.

\bibitem[{Zhao et~al.(2019)Zhao, Peyrard, Liu, Gao, Meyer, and Eger}]{zhao-etal-2019-moverscore}
Wei Zhao, Maxime Peyrard, Fei Liu, Yang Gao, Christian~M. Meyer, and Steffen Eger. 2019.
\newblock \href {https://doi.org/10.18653/v1/D19-1053} {{M}over{S}core: Text generation evaluating with contextualized embeddings and earth mover distance}.
\newblock In \emph{Proceedings of the 2019 Conference on Empirical Methods in Natural Language Processing and the 9th International Joint Conference on Natural Language Processing (EMNLP-IJCNLP)}, pages 563--578, Hong Kong, China. Association for Computational Linguistics.

\bibitem[{Zhao et~al.(2023{\natexlab{a}})Zhao, Strube, and Eger}]{zhao-etal-2023-discoscore}
Wei Zhao, Michael Strube, and Steffen Eger. 2023{\natexlab{a}}.
\newblock \href {https://doi.org/10.18653/v1/2023.eacl-main.278} {{D}isco{S}core: Evaluating text generation with {BERT} and discourse coherence}.
\newblock In \emph{Proceedings of the 17th Conference of the European Chapter of the Association for Computational Linguistics}, pages 3865--3883, Dubrovnik, Croatia. Association for Computational Linguistics.

\bibitem[{Zhao et~al.(2023{\natexlab{b}})Zhao, Ren, Hessel, Cardie, Choi, and Deng}]{zhao2023inthe}
Wenting Zhao, Xiang Ren, Jack Hessel, Claire Cardie, Yejin Choi, and Yuntian Deng. 2023{\natexlab{b}}.
\newblock (inthe) wildchat: 570k chatgpt interaction logs in the wild.
\newblock In \emph{The Twelfth International Conference on Learning Representations}.

\end{thebibliography}

\appendix

\section{Dataset for Reference-based Setting}
\label{dataset}

\begin{itemize}
    \item For the easy-writing task, we use the task of emails, letters, and articles generation as testbeds. Specifically, we extracted 147 relevant instances of email, letter, and article writing from Wildchat \citep{zhao2023inthe}, a real-world user-ChatGPT interactions corpus, as the easy-writing dataset.
    \item For the challenge-writing task, we employ the ``Related Work'' generation task \citep{liu-etal-2023-causal,chen-etal-2021-capturing} as the testbed. In academic manuscripts, the ``Related Work'' sections position the authors' contributions within the existing academic context. They highlight the novelty and advantages of the presented research in comparison to existing works. Specifically, We have randomly selected 90 ``Related Work" paragraphs from scholarly articles within the ACL dataset \citep{rohatgi2023acl} as the challenge-writing dataset.
\end{itemize}

\section{Text Generation Models for Easy Writing Task}
\label{mistral}
We use the APIs of Mistral-7B\footnote{https://huggingface.co/mistralai/Mixtral-8x7B-Instruct-v0.1} and Mistral-8x7B\footnote{https://huggingface.co/mistralai/Mixtral-8x7B-Instruct-v0.1}, hosted on Huggingface, to generate responses for the prompts within our constructed easy-writing task dataset. The parameters for the inference process are shown in Table \ref{tab:easy_task_gen}.

\begin{table}[h]
\centering
\begin{tabular}{ll}
\toprule
\textbf{Parameter}         & \textbf{Value} \\ 
\midrule
temperature       & 0.9           \\ 
max\_new\_tokens   & 4096          \\ 
stop\_sequences   & ["</s>"]      \\ 
top\_p            & 0.95          \\
repetition\_penalty & 1.0           \\ 
do\_sample         & True          \\
seed              & 41            \\
\bottomrule
\end{tabular}
\caption{Generation Configuration Parameters}
\label{tab:easy_task_gen}
\end{table}

\section{Text Generation Models for Challenge Writing Task}
\label{sec:cot}

In academic manuscripts, the ``Related Work'' sections position the authors' contributions within the existing academic context. They highlight the novelty and advantages of the presented research in comparison to existing works. 

In the ``Related Work'' generation task, the model utilizes a set of reference papers, denoted as $Ref$, along with a description of the user's current research denoted as $D$, to generate a ``Realted Work'' draft, denoted as $Y_{Draft}$, for the user. 
In this work, the input data is sourced from the ACL papers \citep{rohatgi2023acl}. We select the related work section paragraph as the test data based on the section title. Notably, the abstracts of both the reference papers and the user's target paper are utilized to encapsulate the core content of the respective research, thereby assisting in the generation process. For both Vanilla GPT-4 and CoT-based GPT-4, the temperature is set as 1.0 in the generation process.

\begin{equation}
    Y_{Draft} = LLM_{gen}(Ref, D)
\end{equation}

\textbf{Vanilla GPT-4}
We concatenate the task instruction and metadata of reference and source papers to directly prompt the LLM to get the final ``Related Work''.

\textbf{CoT-based GPT-4}
We initially prompt the LLM to generate learned relevant knowledge in the training stage\citep{sun2023recitationaugmented} and then create three segments for different perspectives. Finally, these segments, along with the recalled knowledge and the metadata of the input papers, are integrated to generate the comprehensive ``Related Work'' paragraph.
Based on the intermediate step, the LLM can better capture interrelationships among scientific publications and concepts.

\section{Example for Revision Action Item}

\subsection{Revision Regarding the Text Content}

\begin{figure*}[htbp]
    \centering
    \begin{lstlisting}[style=jsonstyle, breaklines=true, breakatwhitespace=true]
{
    "action_name": "simplify",
    "revision_description": "Simplified the text by removing details regarding CDSMs and the inclusion of the SICK benchmark.",
    "revision_level": "reference",
    "revision_intention": "simplify",
    "original_snippet": "Semantic Textual Similarity (STS), the task of measuring the degree of semantic equivalence between two pieces of text, has been extensively explored in the literature. The development of compositional distributional semantic models (CDSMs) forms an integral part of STS investigations, which employ meaning-rich computational systems to better understand and quantify semantic relatedness. The work done by Marelli et al.  went a step further, focusing on producing a large English benchmark, SICK, for the deep evaluation of CDSMs, significantly contributing to the body of tools available for STS analysis.",
    "revised_snippet": "In the broad field of Semantic Textual Similarity (STS), earlier works have explored numerous computational models to comprehend and quantify the semantic relatedness between texts."
}
\end{lstlisting}
    \caption{An example of content-based revision. The generated revision is about simply the background introduction in the AI-generated text.}
    \label{fig:reviseOutput}
\end{figure*}

\textbf{\#\#\#Human-written Text}: 
To deal with the STS task, previous studies have resorted to various features (e.g. word overlap, synonym/antonym), linguistic resources (e.g. WordNet and pre-trained word embeddings) and a wide assortment of learning algorithms (e.g. Support Vector Regression (SVR), regression functions and NNs). Among these works, several techniques extract multiple features of sentences and apply regression functions to estimate these similarity scores (Lai \& Hockenmaier, 2014; Zhao et al., 2014; Bjerva et al., 2014; Severyn et al., 2013). Lai \& Hockenmaier (2014) analyzed distinctive word relations (e.g. synonyms, antonyms, and hyperonyms) with features based on counts of co-occurences with other words and similarities between captions of images. Zhao et al. (2014) predicted the sentence similarity from syntactic relationship, distinctive content similitudes, length and string features. Bjerva et al. (2014) also utilized a regression algorithm to foresee the STS from different features (WordNet, word overlap, and so forth). Finally, Severyn et al. (2013) combined relational syntactic structures with SVR.\\

\textbf{\#\#\#AI-generated Text:} 
Semantic Textual Similarity (STS), the task of measuring the degree of semantic equivalence between two pieces of text, has been extensively explored in the literature. The development of compositional distributional semantic models (CDSMs) forms an integral part of STS investigations, which employ meaning-rich computational systems to better understand and quantify semantic relatedness. The work done by Marelli et al. went a step further, focusing on producing a large English benchmark, SICK, for the deep evaluation of CDSMs, significantly contributing to the body of tools available for STS analysis. $\ldots$.\\

\subsection{Revision Regarding the Text Structure}

\textbf{\#\#\#Human-written Text:} 
First, \textbf{Authors A} proposed a BERT-based method. Second, \textbf{Authors B} proposed a GPT-based method. 

\textbf{\#\#\#AI-generated Text:}
First, \textbf{Authors B} proposed a GPT-based method. Second, \textbf{Authors A} proposed a BERT-based method.

As shown in Table \ref{tab:structure}, current commonly used metrics tend to assign near-perfect scores to AI-generated texts, implying a high degree of equivalence with their human-written counterparts. 
However, this fails to capture the underlying structural differences between the texts. 
As depicted in Figure \ref{fig:structure}, our metric can better capture the structural differences between the texts. Notably, our metric can offer users a coherent and transparent explanation of the scores assigned, benefiting  from the detail of revision actions.

\begin{table}[hbtp]
    \centering
    \begin{tabular}{p{1cm}p{1.5cm}p{1.5cm}p{2cm}}    
    \toprule
    Metric & Rouge-1 & Rouge-2 & Bert-Score\\
    \midrule
    Value & 100.0 & 100.0 & 99.0 \\
    \bottomrule
    \end{tabular}
    \caption{Current typical automated metrics' output. }
    \label{tab:structure}
\end{table}

\begin{figure}[hbtp]
    \centering
    \begin{lstlisting}[style=jsonstyle, breaklines=true, breakatwhitespace=true]
{
 "action_name": "Reorder",
 "revision_description": "Reordered the sequence of references to match the human-written text",
 ...
}\end{lstlisting}
    \caption{In this case, the difference between the two texts is the related work statement order, which represents the author’s argumentation structure}
    \label{fig:structure}
\end{figure}

\section{Revision Categories for the LLM-Generated ``Related Work''}
\label{cat_rev}
\begin{enumerate}
    \item Reference Order Revision: This refers to reorganizing the sequence of references from various viewpoints such as chronological order, methodological approach, or problem context.
    \item Reference Comparison Revision: This refers to integrating comparative discussions among a collection of references, thereby stating their congruities or discrepancies.
    \item Reference Description Revision: This refers to modifying the description of a particular reference paper, either by elaborating it or by making it more concise.
\end{enumerate}


\end{document}